\documentclass[lettersize,journal]{IEEEtran}
\usepackage{amsmath,amsfonts,amsthm}
\usepackage{mathtools}
\usepackage{array}
\usepackage[caption=false,font=normalsize,labelfont=sf,textfont=sf]{subfig}
\usepackage{textcomp}
\usepackage{stfloats}
\usepackage{url}
\usepackage{verbatim}
\usepackage{graphicx}
\usepackage{cite}
\usepackage{bm}
\newtheorem{theorem}{Theorem}
\usepackage{algorithm}
\usepackage{algpseudocode}
\usepackage{siunitx}
\usepackage{caption}
\usepackage{multirow}
\usepackage{amsmath,amsfonts,amssymb,mathrsfs}  
\usepackage{booktabs}
\usepackage{graphicx}
\usepackage{subcaption}
\usepackage[normalem]{ulem}
\usepackage{threeparttable}
\usepackage{multirow}
\usepackage{tabu}
\algtext*{EndWhile}  
\algtext*{EndFor}    

\begin{document}

\title{Provably Safe Trajectory Generation for Manipulators Under Motion and Environmental Uncertainties}

\author{Fei Meng, Zijiang Yang, Xinyu Mao, Haobo Liang, and Max Q.-H. Meng,~\IEEEmembership{Fellow,~IEEE}

\thanks{Manuscript received April 19, 2021; revised August 16, 2021.
This work  was partially funded by Hong Kong ITC-sponsored InnoHK Center (HKCRC) (\textit{Corresponding author: Haobo Liang})}
\thanks{Fei Meng and Haobo Liang are with the Hong Kong Center for Construction Robotics, The Hong Kong University of Science and Technology, Hong Kong, China (e-mail: \{feimeng, hbliang\}@ust.hk)}
\thanks{Zijiang Yang is with the Department of Mechanical and Automation Engineering, The Chinese University of Hong Kong, Hong Kong, China (e-mail: jimmyyang@link.cuhk.edu.hk)}
\thanks{Xinyu Mao is with the Department of Electronic Engineering, The Chinese University of Hong Kong, Hong Kong, China (e-mail: maoxinyu@link.cuhk.edu.hk)}
\thanks{Max Q.-H. Meng is with Shenzhen Key Laboratory of Robotics Perception and Intelligence and the Department of Electronic and Electrical Engineering at Southern University of Science and Technology, Shenzhen, China, and also a Professor Emeritus in the Department of Electronic Engineering at The Chinese University of Hong Kong, Hong Kong, China (e-mail: max.meng@ieee.org)}}

\markboth{Journal of \LaTeX\ Class Files,~Vol.~14, No.~8, August~2021}%
{Shell \MakeLowercase{\textit{et al.}}: A Sample Article Using IEEEtran.cls for IEEE Journals}

\IEEEpubid{0000--0000/00\$00.00~\copyright~2021 IEEE}

\maketitle

\begin{abstract}
Robot manipulators operating in uncertain and non-convex environments present significant challenges for safe and optimal motion planning. Existing methods often struggle to provide efficient and formally certified collision risk guarantees, particularly when dealing with complex geometries and non-Gaussian uncertainties. This article proposes a novel risk-bounded motion planning framework to address this unmet need. Our approach integrates a rigid manipulator deep stochastic Koopman operator (RM-DeSKO) model to robustly predict the robot's state distribution under motion uncertainty. We then introduce an efficient, hierarchical verification method that combines parallelizable physics simulations with sum-of-squares (SOS) programming as a filter for fine-grained, formal certification of collision risk. This method is embedded within a Model Predictive Path Integral (MPPI) controller that uniquely utilizes binary collision information from SOS decomposition to improve its policy. The effectiveness of the proposed framework is validated on two typical robot manipulators through extensive simulations and real-world experiments, including a challenging human-robot collaboration scenario, demonstrating sim-to-real transfer of the learned model and its ability to generate safe and efficient trajectories in complex, uncertain settings.
\end{abstract}

\def\abstractname{Note to Practitioners}
\begin{abstract}
Robot arms working alongside humans must move safely even when their own motions are imprecise and the workspace is cluttered. Existing solutions typically impose overly conservative safety margins, sacrificing efficiency. This paper provides a framework that gives a mathematical guarantee that the robot's collision probability stays below a user-defined threshold, while still allowing efficient motion. A learned predictive model captures how the real robot moves under uncertainty, and a two-stage verification process quickly filters unsafe trajectories before execution. The system replans at 6 Hz, making it responsive enough for human-robot collaboration tasks such as assembly and packaging. A key practical limitation is the requirement to have both an example path for each start-goal pair (for training the model) and the polynomial equations of the obstacles. Future work will focus on reducing this setup cost and improving computational speed for faster replanning.
\end{abstract}


\section{Introduction}
\IEEEPARstart{R}{obot} manipulators are increasingly deployed to address challenges such as labor shortages, worker fatigue, and the need for operation in diverse and hazardous conditions. However, their practical application is often hindered by imperfect perception and significant tracking errors, which pose considerable challenges to their safe and efficient operation. A long-standing problem in robotics is the motion planning of high-dimensional robots, which involves quickly finding a safe and optimal trajectory \cite{lavalle2006planning,zhang2025motion,li2023relevant}.
\IEEEpubidadjcol
To address uncertainty, chance-constrained motion planners have been developed to ensure that constraints are satisfied with a high probability. These planners have been applied to robots with both simplified \cite{axelrod2018Provably,masahiroono2008Iterative,jasour2023Convex,zheng2024CSBRM} and more complex \cite{dai2019Chance} geometric models. For instance, Dawson \textit{et al.} proposed a chance-constrained trajectory generation approach for high-dimensional robots that accounts for observation and motion uncertainty \cite{dawson2023ChanceConstrained}. However, this method is limited by its assumptions of Gaussian uncertainty and convex obstacles. While techniques exist to convexify non-convex obstacles, they can negatively impact the feasibility of the resulting solution. Much of the existing literature focuses on either uncertain obstacles \cite{jasour2023Convex,dawson2020Provably,quintero-pena2021Robust} or uncertainty in the robot's state \cite{luders2010Chancea,vandenberg2011LQGMP,masahiroono2008Iterative,dai2019Chance}. While some recent works have begun to consider non-Gaussian uncertainties and non-convex obstacles \cite{han2022NonGaussian,han2023real,liu2023RADIUS}, these studies are often restricted to mobile robots approximated as a single convex rigid body and involve computationally demanding collision risk checks.
Novel neural networks are proposed to generate risk bounded paths quickly in challenging uncertain nonconvex environments \cite{meng2022nr,meng2023learning}.
However, the generated paths are geometric and require many successful expert paths to learn.
Time-informed set can be built as heuristics based on the deep Koopman operator to accelerate finding the optimal kinodynamic solution; however, it is hard to transfer to robotic manipulators \cite{meng2024online}.

Model Predictive Path Integral (MPPI) control has emerged as a sampling-based stochastic control algorithm for motion planning of high-dimensional robot arms \cite{bhardwaj2021STORM}. While effective for the example manipulators \cite{pezzato2025sampling}, its performance degrades when significant tracking errors in joint space are present. Although some works have been developed to handle system dynamics with neural networks \cite{williams2016information}, propagate uncertainty \cite{mohamed2025towards}, or address parameter uncertainty \cite{abraham2020model}, they often focus on deterministic systems with noisy inputs or are limited to mobile robots in deterministic environments \cite{higgins2023model,yinRiskAwareModelPredictive2022}.

\begin{figure}
\centering
\includegraphics[width=0.49\textwidth]{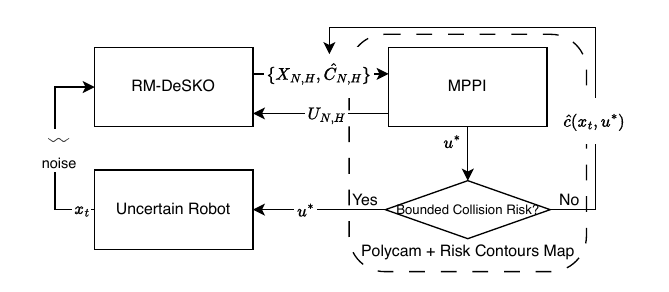}
\caption{Risk-bounded motion planning framework for a manipulator under environmental and motion uncertainties. It combines a RM-DeSKO neural network model predicting states $X_{N,H}$ and costs $\hat{C}_{N,H}$, with a MPPI refining control inputs $u^*$. A supervisory logic enforces environment safety by computing contact force-based collision costs in simulation and checking bounded collision risk for the next predicted state before execution. Noisy state $x_t$ closes the perception-action loop. The new cost function $\hat{c}(x_t, u^*)$ quickly guides stochastic optimization while ensuring risk constraints.}
\label{fig:scheme}
\end{figure}

In this article, we address the unmet need for risk-bounded trajectory generation for uncertain robot manipulators operating in uncertain, non-convex environments. The proposed workflow is illustrated in Fig. \ref{fig:scheme}. Our approach builds upon the success of deep stochastic Koopman operator (DeSKO) in soft manipulator end-effector control \cite{han2022DeSKO}, adapting it to the unique motion planning challenges of rigid manipulators where the state space, objectives, and constraints are fundamentally different. 
Notably, without collecting real data, our neural network trained in simulation generalizes well to real-world scenarios.
 Our primary contribution is a novel framework that integrates a rigid manipulator DeSKO (RM-DeSKO) model for robust state prediction with a hierarchical verification method for efficient and formally certified collision risk assessment. The RM-DeSKO model propagates the distribution of observables in a lifted linear space to predict future states over a planning horizon. We leverage the inferred state distribution and contact force tensors from IsaacGym \cite{pezzato2025sampling} to efficiently compute stable collision costs. To ensure both planning efficiency and formal certification, our hierarchical method uses sum-of-squares (SOS) programming to filter control signals, guaranteeing a bounded collision probability in a risk contours map. Our MPPI-based controller reuses the binary collision results from failed controls to update its policy, leveraging the gradient-free nature of the algorithm.

Our main contributions are summarized as follows:
\begin{itemize}
	\item We formulate and solve, for the first time to our knowledge, the risk-bounded motion planning problem for uncertain robot manipulators facing nonconvex obstacles with probabilistic geometry, size, and location.
	\item We develop a Koopman operator neural network model to predict the states of high-dimensional robot arms under motion uncertainty and develop an efficient, hierarchical collision risk verification method that combines risk contours with formal certification.
	\item We demonstrate the efficiency of our framework through extensive experiments.  We further successfully transfer the policy from simulation to the real robot in a challenging human-robot collaboration scenario. 
\end{itemize}

\section{Preliminaries and Problem Statement}
\subsection{Risk Contours Map}
We denote $W\subset\mathbb{R}^3$ as the robot workspace.
A set of static but uncertain 3D obstacles in the workspace are denoted by $\mathcal{O}=\{\mathcal{O}_1,\dots,\mathcal{O}_{n_o}\}\subset W$, and each one has a polynomial representation in coordinate positions $\bm{p}$
\begin{equation}
\mathcal{O}_i\left(\omega_i\right)=\left\{\bm{p} \in W:o_i\left(\bm{p}, \omega_i\right) \leq 0\right\}, i=1, \cdots, n_o,    
\label{eq:u_obs}
\end{equation}
where $o_i\in\mathbb{R}^4\rightarrow\mathbb{R}$ is the polynomial and ${\omega}_i$ is a random variable with an arbitrary probability distribution provided, not necessarily Gaussian. 
Eq. \eqref{eq:u_obs} can describe convex and nonconvex probabilistic obstacles with uncertain size, location, or geometry.
For instance, {a spherical uncertain obstacle is} ${o(\bm{p},\omega)\leq0}:\{(x,y,z):\,x^2+y^2+z^2-\omega^2\leq0,\,\omega\sim U(1,2)\}$.
Its radius $\omega$ is uncertain and follows a uniform distribution. 

Inflating nonconvex obstacles influences solutions' feasibility in cluttered unstructured environments.
We use the risk contours map $\mathcal{M}$ to describe the continuous acceptable safe subspace \cite{jasourRiskContoursMap2019}.
Risk contours enable analytical moment computation and Cantelli-based hard risk bounds that are valid for any uncertainty distribution.
Given a predefined risk tolerance $\Delta\in[0,1]$, the sets of all relatively safe positions where the probability of collision with any of the obstacle $\operatorname{Prob}(\cdot)$ is no greater than $\Delta$ are represented as follows: 
\begin{equation}
    {\mathcal{C}}_{r_i}^{\Delta}=\{\bm{p}\in W:\operatorname{Prob}(\bm{p}\in\mathcal{O}_i)\leq\Delta\},|_{i=1}^{n_o}.
    \label{eq:probcons}
\end{equation}

Denote $P_{1 i}(\bm{p},\omega_i)\!=\!\mathbb{E}\!\left[o_i^2(\bm{p},\omega_i)\right]$ and $P_{2 i}(\bm{p},\omega_i)\!=\!\mathbb{E}\!\left[o_i(\bm{p},\omega_i)\right]$ as the polynomials in terms of the moments of order 2 and 1 of two specific probabilistic distributions of $\omega_i$, respectively.  
The deterministic inner approximations of $\mathcal{C}_{r_i}^{\Delta}$ using the moment-based approximation method are as follows:
\begin{equation}
	\widehat{\mathcal{C}}_{r_i}^{\Delta}\!\!=\!\!\left\{\!{\bm{p}}\!\in\!{W}\!\!:\!\!\!\,\frac{P_{1 i}(\bm{p},\omega_i)\!\!-\!\!P_{2 i}^2(\bm{p},\omega_i)}{P_{1 i}(\bm{p},\omega_i)}\!\leq\!\Delta,
	P_{2 i}(\bm{p},\omega_i)\!\leq\!0\!
	\right\}\!
	\label{eq:hat contour}
\end{equation}

\subsection{Problem Statement}
This work considers an $n_q$ degrees of freedom serial robotic manipulator with configuration space $Q$. Given a compact time interval $T \subset \mathbb{R}$, we define a trajectory for the configuration as $\bm{q}: T \rightarrow Q \subset \mathbb{R}^{n_q}$ and for the velocity as $\dot{\bm{q}}: T \rightarrow$ $\mathbb{R}^{n_q}$. 
Let the robot state be defined as $\bm{x}_t=[\bm{q}_t, \dot{\bm{q}_t}] \in \mathbb{R}^{2n_q}$ and the commanded action, $\bm{u}_t$, be the desired joint velocity. 
The tracking configurations along the nominal trajectory during execution are subjected to the unknown time-varying process noise $\omega_d\in\mathbb{R}^{2n_q}$.
The dynamical equation of the robot arm under motion uncertainty is
\begin{equation}
    \bm{x}_{t+1}=f(\bm{x}_t,\bm{u}_t)+\omega_d.
    \label{eq:dynamics}
\end{equation}

Given a collision-free start configuration $\bm{q}_0\in Q_{\text{free}}$, a goal region $W_{\text{goal}}\subseteq \{W\setminus\mathcal{O}\}$, and a risk tolerance $\Delta$, we aim to find the actions to steer the end effector of the uncertain manipulator into the $W_{\text{goal}}$ while the probability of collision of its rigid body links $\text{L}(\bm{q}_t)\subseteq W$ is bounded by $\Delta$. 
Thus, the goal is to find a sequence of actions $\{\bm{u}_0,\cdots,\bm{u}_{T-1}\}$ connecting the start and goal to solve the following risk-bounded motion planning problem 
\begin{align}
&\underset{}{\operatorname{min}}
\sum_0^{T-1} C(\bm{x}_t,\bm{u}_t)
\label{eq:prob} \\
\text { s.t. } & \{\bm{q}_0,\bm{q}_T\}\subseteq Q_{\text{free}},\,\text{FK}({\bm{q}}_T)\in W_{\text{goal}}    \\
& \bm{x}_{t+1}=f(\bm{x}_t,\bm{u}_t)+\omega_d \\
\label{eq:limits}
& h_{\lim}({\bm{x}}_t,\bm{u}_t)\leq 0 \\
\label{eq:prob_cons}
& \operatorname{Prob}(\bm{p}_t\in\mathcal{O}_i\mid\forall\bm{p}_t\in\text{L}(\bm{q}_t))\leq\Delta\,|_{i=1}^{n_o}, 
\end{align}
where $\operatorname{Prob}(\cdot)$ is the probability of collision between the arm's volume and uncertain obstacles, $\text{FK}(\cdot)$ is the forward kinematic model, and $h_{\lim}(\cdot)$ denotes the set of inequality constraints, including control, joint position, and velocity limitations.


\section{Receding-Horizon Risk-Bounded Trajectory Generation for Manipulators Under Motion and Environmental Uncertainties}
In this section, we first introduce a sampling-based MPC incorporated with IsaacGym to pave the path of real-time trajectory generation for manipulators. 
Then, to address the misleading costs of rollouts influenced by unknown motion uncertainty, we propose a data-driven method that iteratively predicts the batched states that are most likely to be reached.
In order to guarantee a risk-bounded solution rapidly, we develop a hierarchical collision risk verification approach based on parallelizable physics simulations and sum-of-squares programming.

\subsection{Background on Sampling-based MPC for Manipulators}
Benefiting from parallelizable computation, the sampling-based MPC exhibits remarkable planning efficiency in high-dimensional robot arms.
In general, the MPPI control optimizes time-invariant Gaussian policies $\pi_{\phi_t} = \prod_{h=0}^{H-1} \pi_{\phi_{t,h}}$ over open-loop controls with parameters $\phi_t$, comprising means $\mu_{t}$ and covariances $\Sigma_t$ across a horizon $H$. 
At each iteration, a batch of $N$ control sequences of length $H$, $\bm{U}_{N,H}\in \mathbb{R}^{N\times H\times n_q}$, are sampled. 
B-splines of degree 3 are used to fit controls sampled via a Halton sequence, ensuring smoothness and improved exploration of the action space \cite{bhardwaj2021STORM}. 
Then, people parallelize the rollout of a batch of $N$ future states $\bm{X}_{N,H}\in \mathbb{R}^{N\times H\times 2n_q}$ either by a simple integral model \cite{bhardwaj2021STORM} or the IsaacGym environment \cite{pezzato2025sampling}, and compute associated batched costs $\hat{C}_{N,H}\in \mathbb{R}^{N\times H}$.

The cost function $\hat{c}(\bm{x}_{n,h},\bm{u}_{n,h})$, where $n \in [0,N)$ and $h \in [0,H)$, is structured as a weighted combination of specialized terms, each enforcing a critical aspect of robotic control.
The weights balance these competing objectives, while the individual costs are designed for computational efficiency and parallel evaluation.
The whole state-cost $\hat{C}_n\left(\bm{x}_t, \bm{u}_t\right)$ of a control input sequence is discounted over the planning horizon, i.e., $\sum_{h=0}^{H-1} \gamma^h \hat{c}\left({\bm{x}}_{t, h}, \bm{u}_{t, h}\right)$, where $\gamma\in [0, 1]$ discounts future rewards.
It is involved in the risk-seeking objective function to be minimized by $\phi$.
The mean and covariance of the policy are updated with the following sample-based gradient of weighted sampled control sequences,
\begin{align}
& \Sigma_{t, h}=\left(1\hspace{-0.1cm}-\hspace{-0.1cm}\alpha_\sigma\right) \Sigma_{t-1, h}\hspace{-0.1cm}+\hspace{-0.1cm}\alpha_\sigma \frac{\sum_{i=1}^N w_i\left(u_{t, h}\hspace{-0.1cm}-\hspace{-0.1cm}\mu_{t, h}\right)\left(u_{t, h}\hspace{-0.1cm}-\hspace{-0.1cm}\mu_{t, h}\right)^{\top}}{\sum_{i=1}^N w_i}\nonumber\\
& \mu_{t, h}=\left(1-\alpha_\mu\right) \mu_{t-1, h}+\alpha_\mu \frac{\sum_{i=1}^N w_i u_{t, h}}{\sum_{i=1}^N w_i} 
\end{align}
where $\alpha_{\sigma}$ and $\alpha_{\mu}$ are constants and $w_i$ are importance weights.
Independent adaptation of joint covariances is also enabled to allow optimization of diverse cost terms. 
We execute the first control input from the mean sequence $\bm{u}^*$, then shift the mean and covariance forward while appending default values to warm-start the next optimization cycle.
Interested readers are referred to \cite{bhardwaj2021STORM,pezzato2025sampling} for more details.

\subsection{Deep Stochastic Koopman Operator Model for Uncertain Rigid Manipulator to Propagate in MPPI}
\label{sec:SFO construction}
The accuracy of the forward rollouts plays a key role in the success of MPPI control.
Although the effects of the errors in the unreal model can be overcome to some extent by a receding-horizon manner, a nonlinear transition function with high uncertainty can still cause the algorithm to fail.
Inspired by \cite{han2022DeSKO}, we learn rigid arm dynamics with deep stochastic Koopman operators for kinodynamic motion planning. 
It predict the next states of rigid manipulators under tracking errors, which robustly simulates batched state-control pairs at a low cost.

\begin{figure*}
\centering
\includegraphics[width=0.99\textwidth]{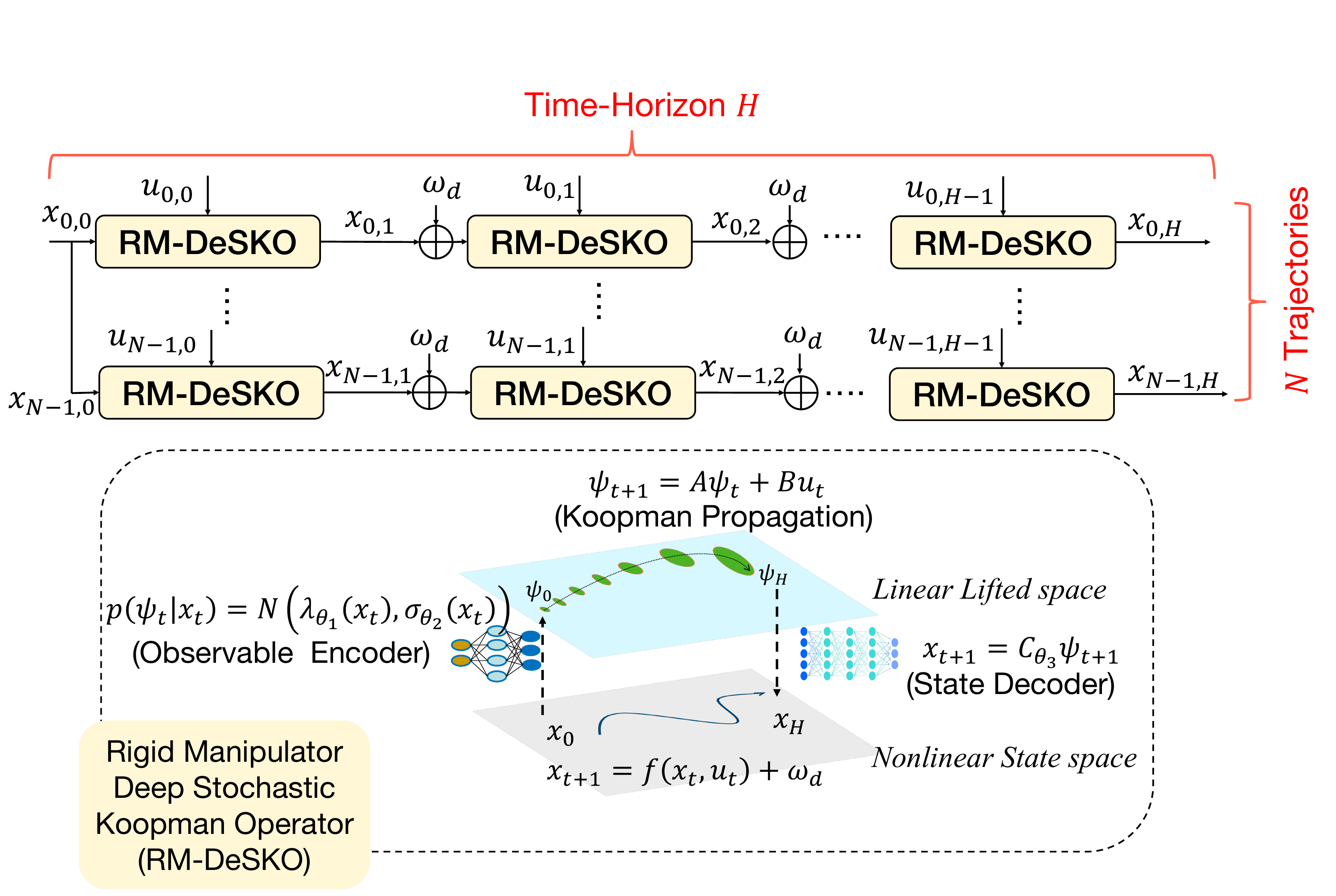}
\caption{Schematic diagram of proposed forward dynamics propagation in MPPI under RM-DeSKO model with key equations for $N$ sampled trajectories over a horizon $H$.}
\label{fig:workflow}
\end{figure*}

We roll out with the RM-DeSKO in the MPPI method to obtain the state evolution of the robot manipulator under uncertainty, as shown in Fig. \ref{fig:workflow}.
Given a batch of $N$ identical current states $\bm{x}_{n,t}$ (where $t=0,\cdots,T$), RM-DeSKO performs iterative parallel propagation using random control signals $\bm{u}_{n,h}$ across the full time horizon $H$.
Here is the underlying mechanism of this network model.
At each time step, it lift the batched noised states $\bm{x}_t$ to have the probabilistic distributions of the observables $\psi$. 
Generally, we assume each observable is subject to a Gaussian distribution.
The mean vector $\lambda_{\theta_1}(\bm{x}_t)$ and diagonal covariance vector $\sigma_{\theta_2}(\bm{x}_t)$ are two encoders parameterized by $\theta_1$ and $\theta_2$, respectively, leading to $\psi_t =\lambda_{\theta_1}\left(\bm{x}_t\right)+\epsilon \sigma_{\theta_2}\left(\bm{x}_t\right),\, \epsilon\sim N(0, I)$.

%

A Koopman operator iteratively propagates the inferred distribution forward, fed with the control inputs, as below. 
\begin{equation}
\psi_{t+k} =A \psi_{t+k-1}+B u_{t+k-1}, k\in[1,H].
\label{eq:KO}
\end{equation}
The Koopman matrix $A$ and control matrix $B$ are represented by the networks of affine linear transformations without biases of suitable sizes.
Finally, we convert the next observables in the lift space to obtain the predicted states by a decoder parameterized by $\theta_3$, i.e., $\bar{\bm{x}}_{t+k}=C_{\theta_3}\psi_{t+k}$.
The RM-DeSKO model is trained to minimize the $H$-step mean squared error loss with a constraint on the entropy of $\psi$.

\subsection{Hierarchical Collision Risk Verification}
\label{sec:collision_risk}
\begin{figure}
\centering
\includegraphics[width=0.4\textwidth]{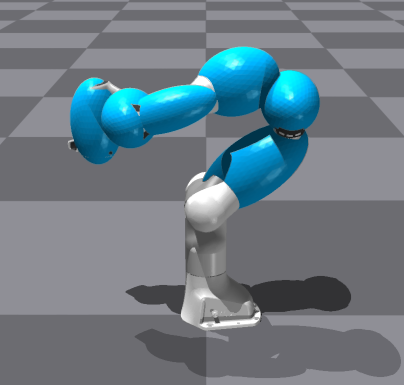}
\caption{Ellipsoidal body description example of a robot arm to be verified for collision risk.}
\label{fig:body}
\end{figure}

We introduce a theorem that verifies whether a tube is inside the risk-bounded region $\hat{\mathcal{C}}_r^{\Delta}(t)$.
\begin{theorem} \text{\cite{jasour2021RealTime}}
A given tube $\mathcal{E}(\mathcal{L}(t))=\left\{\bm{p}\in\mathbb{R}^3:(\bm{p}-\mathcal{L}(t))^TQ(\bm{p}-\mathcal{L}(t))\leq1\right\}$ along a polynomial trajectory $\mathcal{L}(t)$ satisfies the constraint \eqref{eq:prob_cons} over the time horizon $t \in\left[t_0, t_f\right]$ if the polynomials of $\hat{\mathcal{C}}_r^{\Delta}(t)$ take the following SOS representation:
\begin{align}
& P_{2 i}^2\left(\mathcal{L}(t)+\hat{\bm{p}}_0, t\right)-(1-\Delta) P_{1 i}\left(\mathcal{L}(t)+\hat{\bm{p}}_0, t\right)= \sigma_{0 i}\left(t, \hat{\bm{p}}_0\right)\nonumber \\
& +\sigma_{1 i}\left(t, \hat{\bm{p}}_0\right)\left(t-t_0\right)\left(t_f-t\right)+\left.\sigma_{2 i}\left(t, \hat{\bm{p}}_0\right)\left(1-\hat{\bm{p}}_0^T Q \hat{\bm{p}}_0\right)\right|_{i=1} ^{n_o}\!\!, \nonumber\\
& P_{2 i}\left(\mathcal{L}(t)+\hat{\bm{p}}_0, t\right)= \sigma_{3 i}\left(t, \hat{\bm{p}}_0\right)+\sigma_{4 i}\left(t, \hat{\bm{p}}_0\right)\left(t-t_0\right)\left(t_f-t\right)\nonumber\\
& +\left.\sigma_{5 i}\left(t, \hat{\bm{p}}_0\right)\left(1-\hat{\bm{p}}_0^T Q \hat{\bm{p}}_0\right)\right|_{i=1} ^{n_o},
\end{align}
where $Q\in \mathbb{R}^{3\times3}$ is a given positive definite matrix, $\hat{\bm{p}}_0 \in \mathbb{R}^3$ is the variable vector, and $\sigma_{j_i}\left(t, \hat{\bm{p}}_0\right)$, $j=0, \ldots, 5$ are SOS polynomials with appropriate degrees.
\end{theorem}
We assume the robot has state $\bm{q}_t$ during a small time step interval $dt$. 
Given suitable $Q_j$, the movable rigid body links are embraced by consecutive closed ellipsoids whose geometric centers $\bm{p}_t^{c,j}(\bm{q}_t),\,j=1,\cdots,n_q-1$ are obtained by forward kinematics.
Namely, as shown in Fig. \ref{fig:body}, we have $\mathcal{E}(\bm{q}_t)=\bigcup_{j=1}^{n_q-1} \mathcal{E}_j(\bm{q}_t)$, where
\begin{equation}
    \mathcal{E}_j(\bm{q}_t)\hspace{-0.1cm}=\hspace{-0.1cm}\left\{\bm{p}\hspace{-0.05cm}\in\hspace{-0.05cm} \mathbb{R}^3\hspace{-0.1cm}:\hspace{-0.1cm}(\bm{p}\hspace{-0.1cm}-\hspace{-0.1cm}\bm{p}_t^{c,j}(\bm{q}_t))^TQ_j(\bm{p}\hspace{-0.1cm}-\hspace{-0.1cm}\bm{p}_t^{c,j}(\bm{q}_t))\leq1\right\}.
\end{equation}
Similar to the proof in \cite{jasour2021RealTime}, the theorem 1 still holds given $t_0=0$ and $t_f=dt$. 
In other words, considering the hyper-ellipsoids $\mathcal{E}(\bm{q}_t)$ within a small time interval $[t_0,t_f]$ instead of the long tube $\mathcal{E}(\mathcal{L}(t))$ maintains the validity of the theorem.
We can verify the constraint \eqref{eq:prob_cons} using SOS programming for the ellipsoidal volumes at each step.

Ideally, at each iteration, we check the collision risk for the number of $N\times H\times (n_q-1)$ rigid bodies within each planning horizon using SOS programming.
The results are then assigned to collision costs of $\hat{c}(\bm{x}_{n,h},\bm{u}_{n,h})$ for MPPI controller.
It is feasible because they will be binary and MPPI does not need gradients for collision avoidance.
However, a trajectory of length $H=15$ for six ellipsoids even takes about \SI{3.8}{s} on a parallel pool of CPU workers by an off-the-shelf SOS solver, which would be quite computationally expensive for the rollouts.
Furthermore, domain randomization of mesh obstacles in IsaacGym does not support probabilistic geometries and sizes, while sensor noises and limited FOV make the online occupancy map unreliable for the nonconvex obstacles. 
Therefore, we develop a hierarchical collision risk verification method that leverages parallelizable physics simulations based on a high-fidelity 3D scanner and conducts SOS decomposition as a safety filter for the optimized trajectory before execution.

We use Polycam to scan the uncertain nonconvex obstacles and extract their meshes offline.
The collision cost $c_{col}= \alpha\sum F_{contact}$ can be obtained quickly using the contact forces $F_{contact}$ come with IsaacGym \cite{pezzato2025sampling}.
After obtaining an optimized control signal $u^*$ by MPPI, we simulate the model forward one step with our NN model. 
Since the meshes are fixed, we apply the analytical SOS decomposition for the predicted robot arm and polynomial obstacles to determine whether the collision probability is no greater than $\Delta$ before execution.
Thus, we do not need to build the map again when facing probabilistic obstacles. 
We also give stronger certification handling any probability distribution (e.g., uniform, Beta) via moments – non-Gaussian assumption.

The risk-bounded motion planning algorithm for the manipulator under uncertainties is outlined in Alg. \ref{alg:sampling_mpc}.
If any ellipsoid fails to satisfy the SOS condition, which means the probability of collision is greater than $\Delta$ for the next move, we update the corresponding cost and optimize again (Lines 9-12, Alg. \ref{alg:sampling_mpc}).
By combining the contact forces for primary cost computation and the analysis method for falsification, we realize certifiably safe, fast, gradient-free collision risk verification.
Theoretical proof of the bounded collision risk for the manipulator is provided in Appendix \ref{ap:app}.
\begin{algorithm}[t]
\caption{Risk Bounded Motion Generation for Manipulators under Uncertainties}
\begin{algorithmic}[1]
\Require $\bm{q}_0,W_{\text{goal}},H, N, K,\mathcal{M}$
\While{$\bm{p}_{ee}\notin W_{\text{goal}}$}
    \State $\pi_{\bm{q}} \leftarrow shift()$
    \For{$i = 1 \ldots K$} 
        \State $\bm{U}_{N,H} \leftarrow sample\_control(\pi_{\phi_t}, H, N)$
        \State $\bm{X}_{N,H}\leftarrow deep\_Koopman\_operator(\bm{x}_{t}, \bm{U}_{N,H})$
        \State $\hat{C}_{N,H} \leftarrow rollout\_costs\_in\_IsaacGym(\bm{X}_{N,H})$
        \State $\phi_t \leftarrow update\_policy(\hat{C}_{N,H}, \bm{U}_{N,H})$
    \EndFor
    \State $\bm{u}^* = command\_selection(\pi_{\phi_t})$
    \State $\bm{x}_{t+1},\hat{c}(\bm{x}_t,\bm{u}^*)\leftarrow deep\_Koopman\_operator(\bm{x}_t,\bm{u}^*)$
    \While{$\text{NOT SOS\_CONDITION}(\text{L}(\bm{q}_{t+1}),\mathcal{M})$} 
    \State $\phi_t\leftarrow update(\hat{c}(\bm{x}_t,\bm{u}^*),\hat{C}_{N,H}, \bm{U}_{N,H})$
    \State $repeat\_lines\, 8-9$
    \EndWhile
    \State $apply\_control(\bm{u}^*)$
    \State $\bm{x}_t,\,\bm{p}_{ee} \leftarrow read\_state\,\&\,end\_effector\_position()$
\EndWhile
\end{algorithmic}
\label{alg:sampling_mpc}
\end{algorithm}

\section{Experiments}
\subsection{Data Collection and Implementation Details}
We address the risk-bounded motion planning problem for a manipulator without a gripper under environmental and motion uncertainties in both simulation and the real world.
We employ the sampling-based MPC \cite{pezzato2025sampling} as our baseline.
Given a pair of start and goal states in an obstacle-free environment, we collect the nominal trajectory of $l_n$ steps connecting them using the baseline without uncertainties.
We sample $N$ states, $\bm{x}_t$, around each waypoint of the nominal trajectory and propagate the samples for $H=15$ steps with unknown time-varying tracking errors $\omega_d\in\mathbb{R}^{2n_q}$ added at each step.
The actions are collected by sampling Halton splines. 
A dataset containing the number of $l_n\times N\times H$ trajectories under the noise $\omega_d$, $\mathcal{D}=
\left[X_{l_n\times N,\,H} \in \mathbb{R}^{(l_n\times N)\times H\times 2n_q}, U_{l_n\times N,\,T} \in \mathbb{R}^{(l_n\times N)\times H\times n_q}\right]$ is collected.
We configure an Adam optimizer in PyTorch with different learning rates $1e^{-2}$ and $1e^{-3}$ for different parameter groups for training.
The experiments are implemented in Python on Ubuntu 20.04 with an Intel Core i7-10700 CPU processor, 32 GB RAM, and a NVIDIA RTX 4090 GPU.
\footnote{The video recording of simulation and physical experiments is available at \url{https://youtu.be/eAU9YXxO7bA}.}

\subsection{Comparison Results of State Prediction Performance}
We conduct the comparison experiments on a Franka Emika Panda ($n_q=7$) to evaluate the state prediction accuracy for the uncertain manipulator among the well-known neural networks, including Deep Koopman operator with control (DKU) \cite{shi2022Deep}, Long Short-Term Memory (LSTM) \cite{hochreiter1997long}, Multi-Layer Perceptron (MLP), Transformer \cite{vaswani2017attention}, Deep stochastic Koopman operator (RM-DeSKO), and its variants, RM-DeSKO-N and RM-DeSKO-A. 
We set $\bm{x}_t\sim U[-0.2,0.2]\times U[-0.05,0.05]$ and $\omega_d\sim N(0.2,0.2)$ during trajectory sampling.
The hidden layer size of the MLP is $[32,64,128,64,32]$ and activated with ReLU.
The observable function in DKU is parameterized by an encoder with multiple linear layers $[14,128,128,128,128,34]$ concatenated with the state $\bm{x}_t$.
The state and control matrices $A\in\mathbb{R}^{34\times34}$ and $B\in\mathbb{R}^{34\times7}$ are represented as linear layers, respectively.
DKU is trained with the $H$-step mean squared error (MSE) loss.
There are $64$ features in the hidden state and $4$ recurrent layers in the LSTM model.
The Transformer-based model processes historical state trajectories and control inputs to predict future states, using embedding layers to project states and controls into a shared 64-dimensional space, a 2-layer Transformer encoder to capture temporal patterns in the state history, and a multi-head attention mechanism to relate controls to the encoded state memory, finally decoding the attended features through an MLP head to produce $H$-step-ahead predictions of state $x$.
As for RM-DeSKO, the mean vector and the observation matrix, $\lambda_{\theta_1}$ and $C_{\theta_3}$, are parameterized by single-layer linear neural networks with width $64$ followed by a ReLU, respectively, while the covariance vector, $\sigma_{\theta_2}$, has an additional softplus layer.
Different from  RM-DeSKO, the control net of RM-DeSKO-A is parameterized by a single hidden layer with 64 units fed with the state, $B(\bm{x_t})$, and the counterpart of RM-DeSKO-N is $B(\bm{x}_t,\bm{u_t})$. 

\begin{figure}
\centering
\includegraphics[width=0.49\textwidth]{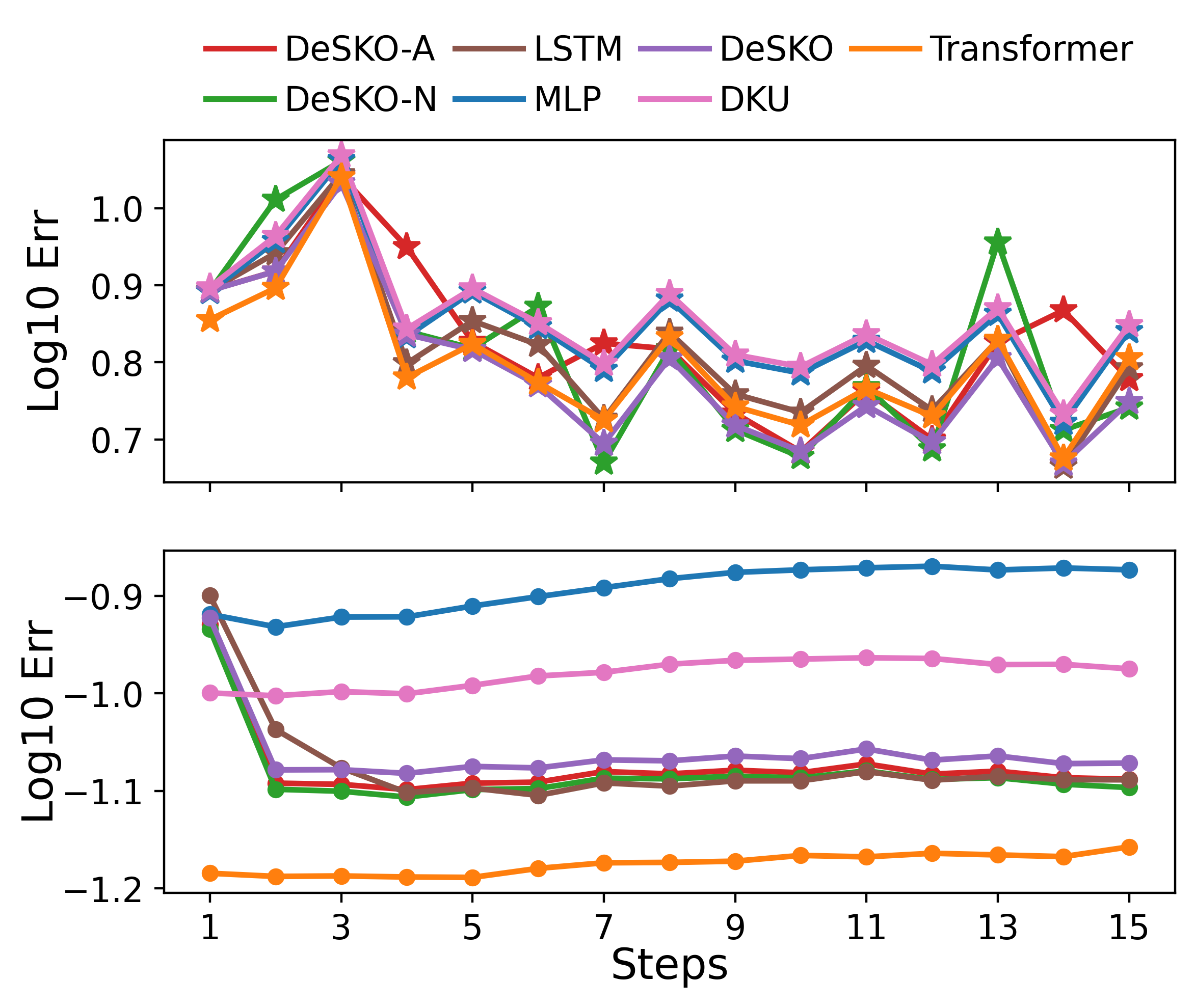}
\caption{Comparison results of the $log10$ of maximum (top) and mean (bottom) prediction errors for a robot arm under state uncertainty across the planning horizon.}
\label{fig:predict}
\end{figure}

We run four times with each of the number of $30k$ trajectories of length $15$ for prediction accuracy evaluation, and Figure \ref{fig:predict} plots the prediction errors across multiple time steps.
Statistical performance over the entire planning horizon is reported in Table \ref{tab:pred_err}.
The Transformer model attains the lowest mean error, while RM-DeSKO generates the lowest maximum error. 
LSTM, RM-DeSKO, RM-DeSKO-N, and RM-DeSKO-A models exhibit similar and slightly higher mean errors. 
DKU and MLP have poor prediction performance due to their simple architectures. 
These results underscore the Transformer’s capability to predict sequential trajectory data effectively, which is attributed to the attention mechanism and hierarchical feature extraction.
The modified RM-DeSKO model also achieves acceptable error reduction given a single-step input. 
The less complex structure also makes it more practical compared with the variants.
The results demonstrate that the RM-DeSKO model is competitive for trajectory prediction, particularly in mitigating unknown noise, which is critical for MPPI control under uncertainty.  

\begin{table*}[t]
\centering
\caption{{Statistical Results of State Prediction Errors Under State Uncertainty Over the Planning Horizon}}
\label{tab:pred_err}
\setlength{\tabcolsep}{5pt}
\begin{tabular}{cccccccc}
    \toprule
    & Transformer & DKU & LSTM & MLP & RM-DeSKO & RM-DeSKO-N & RM-DeSKO-A\\
    \hline
    Max Err& $0.801\pm0.086$&$0.860\pm0.077$ & $0.816\pm0.091$ & $0.852\pm0.078$ & $\bf{0.789\pm0.097}$ & $0.816\pm0.119$ & $0.827\pm0.093$\\
    Mean Err& $\bf{-1.175\pm0.010}$&$-0.980\pm0.014$ & $-1.074\pm0.049$ & $-0.893\pm0.022$ & $-1.061\pm0.037$ & $-1.082\pm0.040$ & $-1.075\pm0.040$\\
    \bottomrule
\end{tabular}
\end{table*}

\subsection{Simulation Experiments for Motion Planning Problem}
In this section, a set containing 10 tasks is tested on the Franka robot in IsaacGym. 
In each task, there are random start and goal configurations and two heart-shaped, uncertain, nonconvex obstacles, whose mathematical expressions are defined as follows:
$(25x_i^2 + \frac{225}{4}y_i^2 + 25z_i^2 - 1)^3 - 3125x_i^2z_i^3 - \frac{5625}{16}y_i^2z_i^3=\omega_i$, where $i=\{1,2\},\,\omega_1\sim U[-0.1,0.1]$ and $\omega_2\sim N(0.05,0.01)$.
Since we can directly generate the 3D meshes for the boundaries of risk contours in simulation, we skip the procedures of Polycam scanning.
The key parameters of MPPI, including the number of samples, tracking error, and the weights of robot-to-goal position, robot orientation, and collision, are set as $N=1000$, $\omega_d\sim N(0.2,0.2)$, and $\alpha_{p}=10$, $\alpha_{o}=0.1$, and $\alpha_{c}=10$, respectively.
The successful distance between the end-effector and the goal is 0.1.
Without loss of generality, we use the end three ellipsoids of the robot for collision risk verification.

\begin{figure*}
 \centering
 \includegraphics[width=0.99\textwidth]{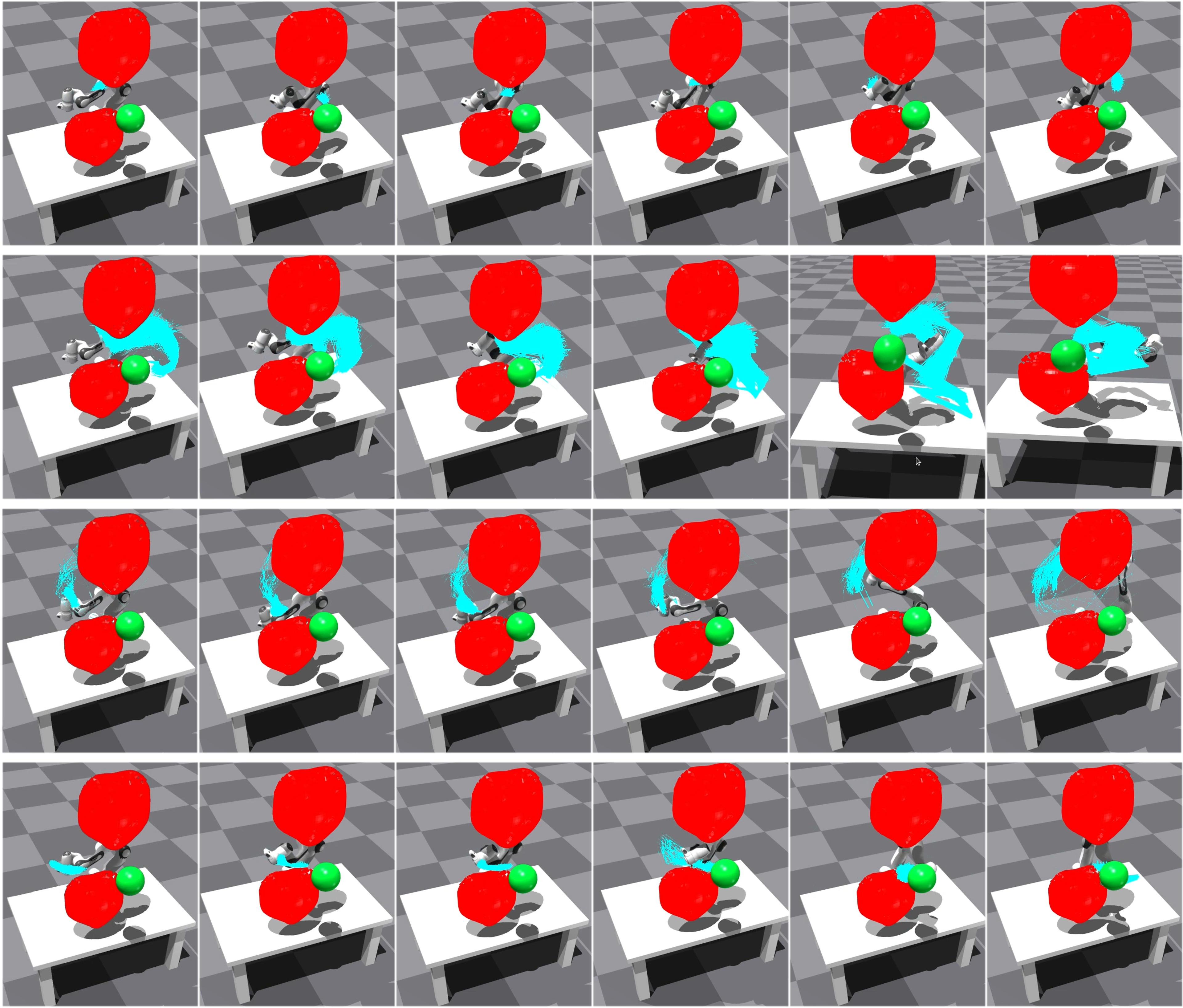}
 \caption{Snapshots of the planned trajectories by the MPPI control using four kinds of nonlinear dynamical systems, i.e., baseline, Transformer, LSTM, and Ours (from top to bottom). The arm tries to reach the goal (green ball) while keeping the probability of collision with uncertain nonconvex obstacles (red heart-shaped) guaranteed. Only our method succeeds in time because of the correct guiding rollouts (cyan) benefited from our neural network dynamical model and collision risk verification.}
 \label{fig:simresults}
 \end{figure*}

We run ten times for each task using Alg. \ref{alg:sampling_mpc} under four dynamics, i.e., the original one in IsaacGym (baseline), Transformer, LSTM, and RM-DeSKO (Ours), and report the success rate, time to goal (TTG), and trajectory length (Length) of the end-effector in Table \ref{tab:simu}.
\begin{table}[tbp]
\centering
\caption{{Comparison Results of Risk-Bounded Motion Planning for Manipulator under Environment and State Uncertainty\label{tab:simu}}}
\begin{tabu}  {X[1,c] X[1.2,c] X[1,c] X[1,c]}
\toprule
Methods & Success Rate $[\%]$ & TTG [s]  & Length \\
\midrule
Baseline & 89 & 47.249  & 2.273\\
Transformer & 0  &  - & -\\
LSTM & 0  &  - & -\\
Ours & 94 &  34.587 & 1.168 \\
\bottomrule
\end{tabu} 
\end{table}
We also present snapshots of executing the planned trajectories in one task under $\Delta=30\%$ in Fig. \ref{fig:simresults}.  
The rollouts of end effector are plotted in cyan, and the meshes of obstacle boundaries are in red.
In the first row, we can see that the guiding rollouts of the baseline are quite transient, causing misleading sample-based gradients of the objective function and wandering in front of the obstacles.
The second row shows that the rollouts based on the Transformer model cannot navigate to the goal.
This is because only the current state and sampled control sequences are fed when rolling out the model at each iteration, whereas the high prediction accuracy for the next possible state depends on the past sequential states as in the last section.
Although we retrain the Transformer with single-step input, it will degrade the effect of the attention mechanism and generate bad rollouts.
As a result, the deviated propagation and policy distribution lead the actions to drive the robot to pass around the obstacles instead of approaching the goal. 
It also fails to compute accurate costs when rolling out based on the LSTM model.
In addition, they require plenty of computational resources, holding them back in MPPI control.
In contrast, our method reaches the goal within the planning time budget with fewer numbers of steps, less time consumed, and a higher success rate than the baseline.
The end effector navigates through the narrow space between the two nonconvex obstacles, as shown in the fourth sub-figure in the bottom row of Fig. \ref{fig:simresults}.
It is the predicted noised states by our neural network that pave the path to update the policy parameters appropriately.

We conduct comparison experiments under different tracking errors, numbers of samples, and pairs of position and orientation cost weights to evaluate the robustness.
In Table \ref{tab:trackingerr}, we can observe that our method scales to the unseen uncertainties.
Since our neural network dynamical model has learned the action trend of the uncertain arm, the rollouts based on it are still useful under the noise outside the domain.
When drawing fewer numbers of samples, our method can consistently find the solution, while the baseline method fails in the first two cases as reported in \ref{tab:numofsample}. 
Furthermore, our method can efficiently find shorter trajectories than the baseline under different pairs of cost weights, as shown in Table \ref{tab:weights}.
However, our method has lower success rates than the baseline when given a low position or a high orientation cost weight.
Since our model is fed by control sequences with reasonable importance sampling weights $w_i(\hat{c}(\cdot))$ during training, the influenced costs $\hat{c}(\cdot)$ make the sequences have a deviation from those in the dataset. 
In short, our approach is robust and efficient for finding risk-bounded trajectories under diverse parameter settings and uncertainties.
\begin{table}[tbp]
\centering
\caption{{Planning Results under Different Tracking Errors $\omega_d$}}
\label{tab:trackingerr}
\begin{tabu}  {X[1.2,c] X[1,c] X[1,c] X[1,c]}
\toprule
Metric& $\mathcal{N}(0.2,0.2)$ & $\mathcal{N}(0.22,0.22)$   &$\mathcal{N}(0.25,0.25)$ \\
\midrule
TTG [s]& 34.587 & 65.992 &100.210 \\
Length  & 1.168 & 1.720  &1.731\\
Success Rate [$\%$]& 94 & 91 & 83\\
\bottomrule
\end{tabu}
\end{table}
\begin{table}[tbp]
\centering
\caption{{Planning Results under Different Numbers of Samples $N$}}
\label{tab:numofsample}
\begin{tabu} {X[1.2,c] X[1,c] X[1,c] X[1,c]}
    \toprule
    Metric & 200  & 500 & 1000 \\
    \midrule
  TTG [s] & 40.529 & 21.147  &34.587 \\
Length & 1.723  & 0.953  &1.168\\
Success Rate [$\%$]& 88 & 93 &94\\
    \bottomrule
\end{tabu}
\end{table}
\begin{table}[tbp]
\centering
\caption{{Planning Results under Different Pairs of Weights\label{tab:weights}}}
\setlength{\tabcolsep}{1pt}
\begin{tabu} to 0.49\textwidth {X[0.8,c] X[1.5,c] X[1,c] X[1,c] X[1,c] X[1,c]}
\toprule
\multirow{3}{*}{Methods} & \multirow{3}{*}{Metric} & \multicolumn{4}{c}{$(\alpha_p,\,\alpha_o)$} \\
\cmidrule(lr){3-6}
& & $(10, 0.1)$ & $(5, 0.1)$ & $(1, 0.1)$ & $(10, 1)$ \\
\midrule
\multirow{3}{*}{Ours} & TTG [s] & 34.587 & 46.839 & 49.586 & 48.325 \\
 &Length & 1.168 & 1.358 & 1.971 & 2.568 \\
 &Success Rate [$\%$]& 94&87&11&8\\
\addlinespace
\multirow{3}{*}{baseline} & TTG [s] & 47.249 & 37.849 & 42.913 & 49.130 \\
 & Length & 2.273 & 2.074 & 2.065 & 2.692 \\
 &Success Rate [$\%$]&89&86&79&71\\
\bottomrule
\end{tabu} 
\end{table}

\subsection{Sim-to-Real Transferred Physical Experiments}
\begin{figure*}[t]
	\centering
	\includegraphics[width=0.98\textwidth]{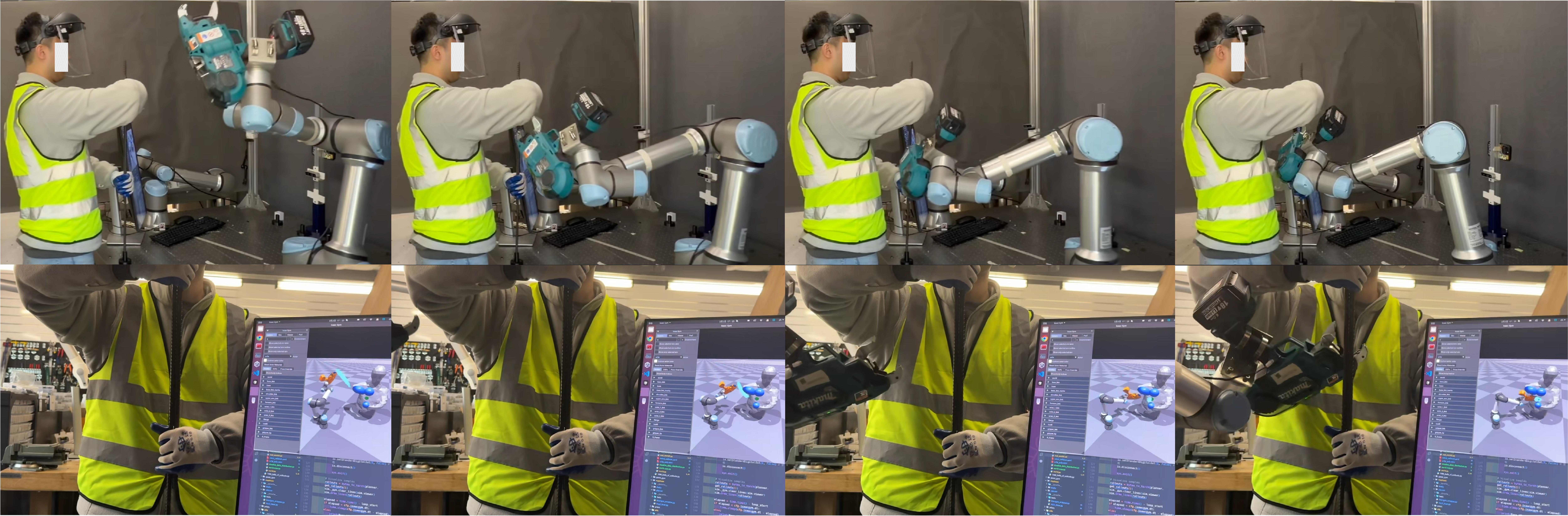} 
	\caption{Snapshots of the rebar tying experiment under uncertainty from the left and front views (top and bottom). The rollouts in cyan on the screen stably point to the goal despite the eccentric payload and tracking errors of the manipulator. The robot successfully generates a risk-bounded trajectory to bypass the uncertain nonconvex human arms and reach the tying target using our method, which transfers directly from simulation to a real robot. }
	\label{fig:physical_experiment}
\end{figure*}

In this section, we validate our method in a human-robot collaboration task where a UR5e manipulator ($n_q=6$) approaches and ties the two vertical rebars held by a sole worker with a rebar tying tool (Makita DTR181) fixed at the end-effector, as shown in Fig. \ref{fig:physical_experiment}.
Since it is hard to construct a complete occupancy map online for this nonconvex environment under the third-view camera's FOV constraints, one scans the worker's body via the Polycam app and extract the corresponding scene collision meshes for collision costs computation.
The obstacles, horizontal human arms, have probabilistic locations due to fatigue after mapping. 
The uncertain and eccentric tying machine payload ([2.7,3.2]kg, caused by consumed tie wires) produces joint state noise. 
Therefore, we need to plan a risk-bounded trajectory that accounts for motion and environmental uncertainties.
Different tracking errors $\omega_d\sim{N}(0,\sigma)$, where $\sigma=\{0.15, 0.20, 0.25\}$, are added to the robot state observations to test our robustness before physical experiments.
Along the nominal trajectory of length $l_n=177$ and range of $\bm{x}_t\sim U[-0.1,0.1]$ around its waypoints, we collect the dataset with $N=400$ batched noise $\omega_d$ at each step in the simulation environment to train the NN model.
We construct several ellipsoids whose centers are subjected to normal distribution which account for the possible occupation area of arms for collision risk verification under $\alpha_{c}=0.01$.
$\lambda_{\theta_1}$ and $\sigma_{\theta_2}$ of RM-DeSKO are parameterized by $[64,128,64]$ linear layers and the other settings are as the same as those for the simulated Franka arm.
We only verify the collision risk between the tying gun and the human arm ellipsoids.
The objective is to let the nozzle of the tying tool arrive the place to be spliced between the hands under a guaranteed collision risk $\Delta=10\%$, and the tying gun automatically binds the two rebars together.

Given a limited time budget, the baseline fails to converge to the target across all three noise levels due to high noised state variance. 
In contrast, our method achieves a consistent 90\% success rate (9 out of 10 trials) under the noises tested, and it takes \SI{8.32}{s}, \SI{8.46}{s}, and \SI{10.21}{s} on average, respectively.
The failure is caused by the consistent disturbance if the small tying tool nozzle misses the tying position of slender reinforcing bars.

Finally, we deploy our method with the same weight file trained before on the real UR5e. 
We keep manually added $\omega_d\sim{N}(0,0.15)$ at each step and the control loop operated at a frequency of \SI{6}{Hz}. 
Just like the simulation results, the baseline method failed to reach the target.
We present a successful execution of our method from front and left views in Fig. \ref{fig:physical_experiment}.
It can be observed that the robot continuously adjusts its configuration to avoid the worker's arms under an acceptable risk tolerance, and ultimately reach the goal and tie.
As shown on the screen, the stable rollouts in cyan predicted by our RM-DeSKO support accurate computation costs and safe trajectory generation. 
The superior performance demonstrates the zero-shot sim-to-real transfer and planning efficiency.



\section{Conclusion}
We presented a receding-horizon motion planning method that guarantees a bounded collision probability for the uncertain manipulators in uncertain nonconvex environments.
Our developed neural network model and hierarchical collision risk verification method significantly improve the obstacle avoidance performance of the MPPI control for the robot arm suffering from joint position and velocity noises, and uncertain surrounding objects.
Scalability in different settings and goal arrival performance with theoretical safety guarantees, shown in the experiment results, validates the robustness and effectiveness.
The Gaussian splatting renderer provides a photorealistic view from any viewpoint. This can be leveraged for perception-aware planning and dynamic risk contour integration for different kinds of obstacles in the future.


{\appendix[Theoretical Guarantees of Bounded Risk for Manipulators under Uncertainties]
\label{ap:app}
Since we infer the joint positions $\bm{q}_t$ using a deep neural network model, the ellipsoids $\mathcal{E}$ built based on them introduce a probability that any part of the forward occupancy $\bm{p}_t\in\text{L}(\bm{q}_t)$ stays inside the ellipsoid.
It is greater than $(1-\Delta_{ell})$, where $\Delta_{ell}$ is usually a quite small constant, e.g., $0.001$.
Mathematically, for $t\in[0,T]$, we have
\begin{equation}
    \operatorname{Prob}\left(\bm{p}_t\in \mathcal{E}(\bm{q}_t)\right)\geq1-\Delta_{ell}.
\end{equation}

We rely on risk contours to verify the collision risk for the rigid bodies in uncertain environments. 
If every ellipsoid is inside the risk-bounded set $\hat{\mathcal{C}}_r^{\Delta}$, then the collision risk constraint \eqref{eq:prob_cons} becomes 
\begin{equation}
\operatorname{Prob}\left(\bm{p}_t \in \mathcal{O} \mid \bm{p}_t\in \mathcal{E}(\bm{q}_t) \subseteq \hat{\mathcal{C}}_r^{\Delta}\right) \leq \Delta_o.
\end{equation}

We optimize configurations $\bm{q}_t$ that make the total cost as small as possible during online MPC optimization. 
We set a high weight $\alpha_c$ to ensure the links are inside  $\hat{\mathcal{C}}_{\mathrm{r}}^{\Delta}$, leading the probability of collision between the robot and any obstacle to be bounded by $\Delta_o+\Delta_{ell}$ as
\begin{equation}
\begin{aligned}
& \operatorname{Prob}\left(\bm{p}_t\in\mathcal{O} \mid \mathcal{E}(\bm{q}_t) \subseteq \hat{\mathcal{C}}_r^{\Delta}\right) \\
& \leq \operatorname{Prob}\left(\bm{p}_t\in\mathcal{O} \mid \bm{p}_t\in \mathcal{E}(\bm{q}_t)\subseteq\hat{\mathcal{C}}_r^{\Delta}\right) \\
& \times\operatorname{Prob}\left(\bm{p}_t\in\mathcal{E}(\bm{q}_t)\right) +\operatorname{Prob}\left(\bm{p}_t \notin\mathcal{E}(\bm{q}_t)\right) \\
& \leq \Delta_o+\Delta_{ell}.
\label{eq:total_prob}
\end{aligned}
\end{equation}

The bound in \eqref{eq:total_prob} applies to the ellipsoids at one planning iteration. 
Suppose the MPC needs to take $z=\frac{T}{dt}$ steps to complete the final solution. 
Therefore, we can derive the probability of the ellipsoids embracing the arm over all the planning iterations as follows
\begin{equation}
\operatorname{Prob}\left(\bm{p}_t\in\mathcal{E}({\bm{q}_t})\right)^z \geq\left(1-\Delta_{ell}\right)^z.
\end{equation}
As \eqref{eq:total_prob}, the probability of collision over all the planning cycles is no greater than
\begin{equation}
\Delta_o+\left(1-\left(1-\Delta_{ell}\right)^z\right)\leq \Delta_o+z\Delta_{ell}.
\end{equation}

Given the user-defined risk tolerance $\Delta$ and the number of $\overline{z}$ 
planning cycles of upper bound guess, we can determine $\Delta_o$ and $\Delta_{ell}$ to have
\begin{equation}
\Delta_o+\overline{z}\Delta_{ell} \leq \Delta.  
\end{equation}
The probability of collision between the uncertain arm and any uncertain nonconvex obstacle over the entire planning horizon $\operatorname{Prob}$ is bounded by $\Delta$ as below
\begin{equation}
\operatorname{Prob}\leq \Delta_o+z\Delta_{ell} \leq \Delta_o+\overline{z}\Delta_{ell} \leq \Delta,
\end{equation}
where $z\leq\overline{z}$ is the actual number of planning iterations.


\bibliographystyle{IEEEtran}
\bibliography{references}

\vfill

\end{document}